\documentclass[conference, letterpaper]{IEEEtran}

\usepackage{amssymb}
\usepackage{algorithm}
\usepackage{algorithmic}
\usepackage{setspace}
\usepackage{flushend}

\usepackage{subcaption}

\ifCLASSINFOpdf
   \usepackage[pdftex]{graphicx}
\else
\fi

\usepackage[cmex10]{amsmath}
\usepackage{color}
\usepackage{fancyhdr}

\renewcommand{\thispagestyle}[2]{} 

\fancypagestyle{plain}{
        \fancyhead{}
        \fancyhead[C]{first page center header}
        \fancyfoot{}
        \fancyfoot[C]{first page center footer}
}
\begin{document}

\title{CUSBoost: Cluster-based Under-sampling with Boosting for Imbalanced Classification}

\author{
\IEEEauthorblockN{Farshid Rayhan, Sajid Ahmed, Asif Mahbub, Md. Rafsan Jani, \\Swakkhar Shatabda, and Dewan Md. Farid}
\IEEEauthorblockA{Department of Computer Science \& Engineering, United International University, Bangladesh\\  
Email: dewanfarid@cse.uiu.ac.bd}}

\maketitle

\begin{abstract}
Class imbalance classification is a challenging research problem in data mining and machine learning, as most of the real-life datasets are often imbalanced in nature. Existing learning algorithms maximise the classification accuracy by correctly classifying the majority class, but misclassify the minority class. However, the minority class instances are representing the concept with greater interest than the majority class instances in real-life applications. Recently, several techniques based on sampling methods (under-sampling of the majority class and over-sampling the minority class), cost-sensitive learning methods, and ensemble learning have been used in the literature for classifying imbalanced datasets. In this paper, we introduce a new clustering-based under-sampling approach with boosting (AdaBoost) algorithm, called CUSBoost, for effective imbalanced classification. The proposed algorithm provides an alternative to RUSBoost (random under-sampling with AdaBoost) and SMOTEBoost (synthetic minority over-sampling with AdaBoost) algorithms. We evaluated the performance of CUSBoost algorithm with the state-of-the-art methods based on ensemble learning like AdaBoost, RUSBoost, SMOTEBoost on 13 imbalance binary and multi-class datasets with various imbalance ratios. The experimental results show that the CUSBoost is a promising and effective approach for dealing with highly imbalanced datasets.     
\end{abstract}


\IEEEpeerreviewmaketitle

\section{Introduction}
In machine learning (ML) for data mining (DM) applications, supervised learning (or classification) is the process of identifying new/ unknown instances employing classifiers (or classification algorithms) based on a group of instances with known class membership (training data) \cite{Farid_ESWA_vol64, Farid_ESWA_vol41, Farid_ESWA_vol40, Farid_IJDMKMP}. Often real-world data sets are multi-class, high-dimensional and class-imbalanced, which fall-off the classification accuracy of many ML algorithms. Therefore, number of ensemble classifiers with sampling techniques have been proposed for classifying  binary-class low-dimensional imbalanced data in the last decade \cite{He, Yanmin-Sun, Wasikowski}. Ensemble classifiers use multiple ML algorithms to improve the performance of individual classifiers that combine multiple hypotheses to form an advance hypothesis \cite{Farid_ESWA_vol40}. The sampling methods use under-sampling (under-sampling of the majority class instances) and over-sampling (over-sampling the minority class instances) techniques to alter the original class distribution of imbalanced data. The under-sampling methods with random sampling of the majority class might suffer from the loss of potentially useful training instances. On the other hand, over-sampling with replacement doesn't significantly improve minority class recognition and increase the likelihood of overfitting \cite{Sun}.  

In real-world class imbalance data sets, the minority class instances are outnumbered by the majority class instances. However, the minority class instances are representing the concept with greater interest than the majority class instances \cite{Farid_Benelearn}. The traditional ML for DM algorithms, such as decision tree (DT) \cite{Farid_ESWA_vol64, Farid_ESWA_vol40}, na\"\i ve Bayes (NB) classifier \cite{Farid_ESWA_vol41}, and k-nearest neighbors (kNN) \cite{Farid_ESWA_vol64}, build the classification models that maximise the classification rate, but ignore the minority class. The most approved methods for dealing with the class imbalance problems are sampling techniques, ensemble methods, and cost-sensitive learning methods. The sampling techniques (under-sampling and over-sampling) either remove the majority class instances from the imbalanced data or add the minority class instances into the imbalanced data to get the balanced data. The ensemble methods such as Bagging and Boosting are also widely used for classifying imbalanced data. Basically, the ensemble methods use sampling technique in each iteration. The cost-sensitive learning is also applied for solving the class imbalance problems, which assigns different cost of misclassification errors for different classes. Usually, high cost for the minority class and low cost for the majority class. But, the classification results are not stable in cost-sensitive learning methods, as it is difficult to get the accurate misclassification cost and different misclassification cost might result in different inductions.    

The methods for dealing with class imbalance problems are divided into two categories: (a) external methods and (b) internal methods. The external methods are also known as data balancing methods, which preprocess the imbalanced data to get the balanced data. The internal methods modify the existing learning algorithms for reducing their sensitiveness to the class imbalance when learning from the imbalanced data. In this paper, we present a new clustering-based under-sampling approach with boosting (AdaBoost), called CUSBoost algorithm. We divide the imbalanced dataset into two part: majority class instances and minority class instances. Then, we cluster the majority class instances into several clusters using k-means clustering algorithm and select the majority class instances from each cluster to form a balanced dataset, where the majority and minority class instances are almost equal. Clustering helps us to group the majority class instances in such a way that instances in the same cluster are more similar to each other than to those in other clusters. So, instead of randomly removing the majority class instances we used clustering technique to select the majority class instances. CUSBoost combines the sampling and boosting methods to form an efficient and effective algorithm for class imbalance learning. We tested the performance of CUSBoost algorithm with AdaBoost, RUSBoost, and SMOTEBoost algorithms on 13 imbalance datasets. Based on the experimental results, we can validate that combining clustering-based under-sampling approach with AdaBoost algorithm is a promising technique for alleviating class imbalance problem.            

The remainder of this paper is organised as follows. Section \ref{ralated_work} presents related works. Section \ref{methods} describes the data balancing methods, and Section \ref{CUSBoost} presents the CUSBoost algorithm. Section \ref{experiments} provides the experimental results. Finally, we conclude in Section \ref{conclusion}. 

\section{Related work}
\label{ralated_work}
\begin{figure}[h]
\centering
\vspace{5mm}
\includegraphics [scale=0.45]{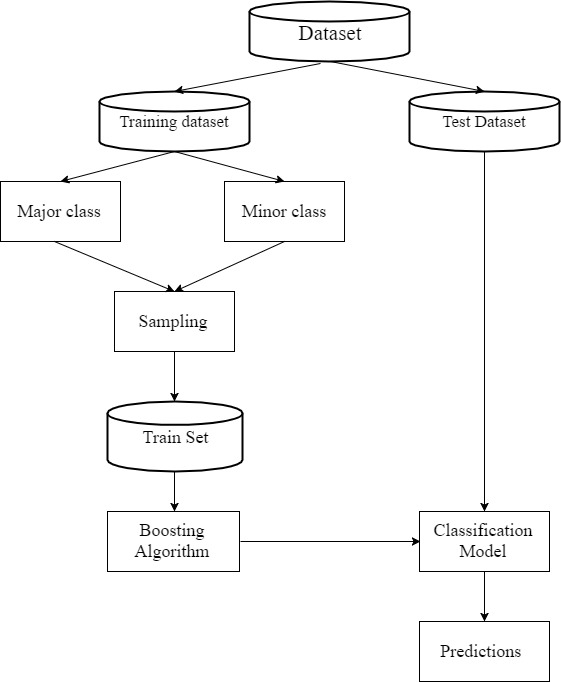}
\vspace{1mm}
\caption{Sampling with boosting for classifying imbalanced data.}
\label{sampling_Boosting}
\end{figure}

In the last decade, sampling methods, bagging and boosting based ensemble methods, and cost-sensitive learning methods, have been used to deal with the imbalanced binary classification problems \cite{Beyan, Pozzolo}. Fig. \ref{sampling_Boosting} shows the process of classifying imbalanced data using sampling with boosting approach. 

Sun et al. \cite{Sun} proposed an ensemble method for addressing binary-class imbalance problems by converting an imbalanced binary learning process into multiple balanced learning processes. In this method, the majority class instances were divided into several groups/ sub-data sets, where each sub-set has the similar number of minority class instances. So, several balanced data sets were generated. Then, each balanced data set was employed to build a binary classifier. Finally, these binary classifiers were combined into an ensemble classifier to classify new data.   

Chawla et al. \cite{Chawla} proposed an over-sampling approach called SMOTE (Synthetic Minority Over-sampling TEchnique) algorithm in which the minority class is over-sampled by creating synthetic minority class instances rather than by over-sampling with replacement. SMOTE generated synthetic instances by operating in \textit{feature space} rather than \textit{data space} employing $k$ minority class nearest neighbors. Their result showed that the combination of over-sampling with under-sampling performed better in Receiver Operating Characteristic (ROC) space. Santos et al. \cite{Santos} implemented a cluster-based (k-means) over-sampling approach where SMOTE was adapted to oversample clusters with reduced sizes. This work considered merging the minority class instances from the multiple over-sampled datasets. Blagus and Lusa \cite{Blagus} investigated the behavior of SMOTE on high-dimensional imbalanced data, where the number of features greatly exceeds the number of training instances. They found that feature selection is necessary for SMOTE with k-nearest neighbors (kNN), as SMOTE strongly biases the classifier towards the minority class.  

Seiffert et al. \cite{Seiffert} presented a new hybrid sampling/boosting algorithm, called RUSBoost, which applied random under-sampling (RUS) with AdaBoost algorithm. RUS randomly removes the majority class instances to form a balanced data. RUSBoost was built based on the SMOTEBoost (synthetic minority over-sampling with AdaBoost) algorithm \cite{Chawla_SMOTEBoost}. SMOTEBoost was built upon over-sampling approach with AdaBoost algorithm. Galar et al. \cite{Galar} presented an ensemble algorithm by evolutionary under-sampling (EUA) approach, called EUSBoost, to classify highly imbalanced datasets. EUS generated several sub-datasets using randomly under-sampling technique and obtained a best under-sampled dataset of the original dataset that cannot be improve further. RUSBoost, SMOTEBoost, and EUSBoost applied data sampling techniques into the AdaBoost.M2 algorithm by considering the minority class instances and the majority class instances. Yen and Lee \cite{Yen} presented a cluster-based under-sampling approach to cluster all the training instances (majority class instances and minority class instances) into some clusters. This approach selected a suitable number of majority class instances from each cluster by considering the ratio of the number of majority class instances to the number of minority class instances in the cluster. 

\section{Data Balancing Methods}
\label{methods}
Researchers have previously proposed various ways of dealing with imbalanced datasets. The following section describes a few of those methods.

\subsection{Sampling Techniques}

\subsubsection{Under-sampling}
\begin{figure}[h]
\centering
\vspace{5mm}
\includegraphics [scale=0.7]{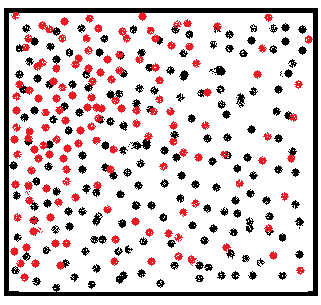}

\vspace{2mm}
\caption{Random under-sampling (RUS) approach. Here the red dots are selected instances from the majority class, where black and red dots are representing all the majority class instances.}
\label{RUS}
\end{figure}

Under-sampling involves the removal of some majority class instances to result in a balanced distribution of all classes \cite{Farid_Benelearn}. However, this can result in the removal of informative instances from the majority class especially if the number of instances of the minority class is very small (for example majority to minority ratio is 100:1). Thus, there will be a huge loss of information from the majority class instances, leading to a non-optimal classification. Fig. \ref{RUS} shows the random under-sampling process to select the majority class instances randomly. 

\subsubsection{Oversampling}
In case of oversampling, we increase the number of  minority class instances and this can be done in different ways. The most common of these methods is random oversampling, where minority class instances are duplicated until both class instances are balanced. The issue with this method, however, is that there is a high probability of over-fitting due to the same instances occurring multiple times. SMOTE (Synthetic Minority class Oversampling TechniquE) is another oversampling technique which handles this problem by creating synthetic instances of the minority class instead of duplicating the minority class instances \cite{Chawla}. This is done by by interpolating several minority class instances that lie together. Nevertheless, this problem can be present to some extent because the newly generated samples could be almost identical to the existing minority class instances. Modified synthetic minority oversampling technique (MSMOTE) is a modified version of SMOTE that divides the instances of the minority class into three groups: safe, border, and latent noise instances. This is done by the calculation of the distances among all instances.

Under-sampling method generally works better than oversampling methods as long as the imbalance ratio of the dataset is not very high \cite{Farid_Benelearn}.

\subsection{Cost-sensitive learning}
This method tackles the class imbalance problem by assigning more misclassification cost to minority class instances for the underlying classifier. The misclassification cost can be used in the cost function and the cost function can be optimised by the classifier. The classifier will now treat both majority and minority classes equally due to their underlying costs. But in this case, the greatest challenge is in assigning the costs as they are difficult to be derived from datasets. 

\subsection{Ensemble Learning} 
Ensemble learning methods combine multiple base learners which may be the same or different. This usually increases the predictive capability of the individual base classifiers, thus making it adaptive to different datasets. Ensemble methods find their usage in a wide variety of problems.

\subsubsection{Bagging} 
Bagging creates multiple sub-sets of the original dataset through sampling with replacement or without replacement. These sub-sets are used by the base learners while considering each instance with equal weight. The output of the individual base learners are considered as votes to determine the final prediction.

\subsubsection{Boosting} 
Boosting is similar to bagging in that it combines multiple base learners to obtain a result based on voting technique. However, it's differences lie in that boosting assigns weight to instances according to how hard they are to classify, thus setting high weight of instances that are difficult to classify. One base learner contributes to the weight used by the next base learner. Weights are assigned to the base learners as well according to their predictive accuracy and this is taken into consideration when classifying a new test instance. Although boosting was not created specifically for imbalanced dataset problems, it's characteristic of assigning weights to examples that are relatively harder to classify makes it ideal for dealing with class imbalance problems. It can be somewhat referred to as a cost-sensitive method.

\section{CUSBoost Algorithm}
\label{CUSBoost}
CUSBoost is based on the combination of cluster-based sampling and Adaboost algorithm. It is similar to RUSBoost and SMOTEBoost with the critical difference occurring in the sampling technique. SMOTEBoost uses SMOTE method to oversample the minority class instances, while RUSBoost uses random under-sampling on the majority class. In comparison, our proposed CUSBoost uses cluster-based sampling from the majority class. CUSBoost separates the majority and minority class instances from the original dataset and clusters the majority class instances into $k$ clusters using k-means clustering algorithm. Here, the parameter $k$ is determined by hyper-parameter optimisation. After that, random under-sampling is performed on each of the created clusters by randomly selecting 50\% of the instances (but it can be tuned according to the domain problem or dataset) and removing the rest. As clustering is used before sampling, theoretically this algorithm will perform best when the dataset is highly cluster-able. These representative samples are then combined with the minority class instances to obtain balanced datasets. Our algorithm's strength lies in the fact that it considers examples from all subspaces of the majority class, since k-means clustering puts each instance in some cluster. Other similar methods often fail in obtaining proper representatives of the majority class. Fig. \ref{CUSBoost} shows the proposed cluster-based under-sampling technique to select the majority class instances.

\begin{figure}[h]
\centering
\vspace{5mm}
\includegraphics [scale=0.7]{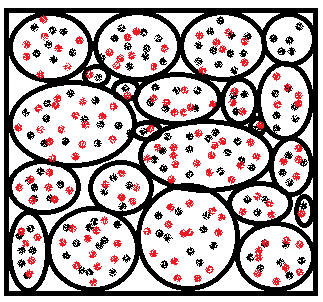}
\vspace{2mm}
\caption{Cluster-based under-sampling (CUS) approach. Here the red dots are selected instances from the majority class, where black and red dots are representing all the majority class instances.}
\label{CUSBoost}
\end{figure}

CUSBoost considers a series of decision trees using C4.5 algorithm and combines the votes of each individual tree to classify new instances. Initially, each instance is initialised with an equal weight, $\frac {1} {d}$, where $d$ is the total number of training instances. The weights of instances are adjusted according to how they were classified. If an instance was correctly classified then its weight is decreased, or if misclassified then its weight is increased. The weight of an instance reflects how difficult it is to classify. To compute the \textit{error rate} of model $M_{i}$, we sum the weights of misclassified instances in $D_{i}$ that is shown in Eq. \ref{IJCA_p1_equ5}. If an instance, $x_{i}$ is misclassified, then $err(x_{i})$ is one. Otherwise, $err(x_{i})$ is zero (when $x_{i}$ is correctly classified).

\begin{equation}
error(M_{i}) = \sum_{i=1}^{d} w_{i} * err(x_{i})
\label{IJCA_p1_equ5}
\end{equation}

If an instance, $x_{i}$ in $i$th iteration is correctly classified, it's weight is multiplied by error $(\frac {error(M_{i})} {1-error(M_{i})})$. Then the weights of all instances (including misclassified instances) are normalised. To normalise a weight, we multiply it by the sum of old weights, divided by the sum of new weights. As a result, the weights of misclassified instances are increased and the weights of correctly classified instances are decreased. If the \textit{error rate} of model $M_{i}$ exceeds 0.5 then we abandon $M_{i}$ and derive a new $M_{i}$ by generating a new sub-data set $D_{i}$. The CUSBoost algorithm is summarised in Algorithm \ref{CUSBoost_Algo}.

\begin{algorithm}
   \caption{CUSBoost Algorithm} \label{CUSBoost_Algo}
   \textbf{Input:} Imbalanced data, $D$, number of iterations, $k$, and C4.5 decision tree induction algorithm. \\
   \textbf{Output:} An ensemble model. \\  
   \textbf{Method:}
   \begin{algorithmic}[1]
	\STATE initialize weight, $x_{i} \in D$ to $\frac {1} {d}$;
        \FOR{i = 1 to k }
       	    \STATE create balanced dataset $D_{i}$ with distribution $D$ using cluster-based under-sampling;   
       	    \STATE derive a tree, $M_{i}$ from $D_{i}$ employing C4.5 algorithm;
       	    \STATE compute the error rate of $M_{i}$, $error(M_{i})$;
	    \IF{$error(M_{i})$ $\geq$  0.5}
		\STATE go back to step 3 and try again;
	    \ENDIF
	    \FOR{each $x_{i} \in D_{i}$ that correctly classified}
	    	\STATE multiply weight of $x_{i}$ by $(\frac {error(M_{i})} {1-error(M_{i})})$; // update weights
	    \ENDFOR
	    \STATE normalise the weight of each instances, $x_{i}$;
	\ENDFOR
   \end{algorithmic} 
   To use the ensemble to classify instance, $x_{New}$: \\   
   \begin{algorithmic}[1]
	\STATE initialise weight of each class to 0;
        \FOR{i = 1 to k} 
       	    \STATE $w_{i} = log \frac{1-error(M_{i})}{error(M_{i})}$; // weight of the classifier's vote   
       	    \STATE $c = M_{i}(x_{New})$; // class prediction by $M_{i}$ 
       	    \STATE add $w_{i}$ to weight for class $c$;
	\ENDFOR
	\STATE return the class with largest weight;
   \end{algorithmic}
\end{algorithm}
    
\section{Experimental Results}
\label{experiments}
In this section, we present the experimental analysis to examine the performance of our proposed CUSBoost algorithm. We have used datasets from KEEL-dataset repository \cite{KEEL-dataset} with different imbalance ratio. Table \ref{datasets} show the datasets details . 
 
\begin{table}[h!]\footnotesize
  \centering
  \caption{Imbalanced datasets description.}\label{datasets}
  \begin{tabular}{|l|l|l|l|l|l|l|}	\hline
        	No.	& Datasets						& Instances	& Features	& Class		& Imbalance  	\\
		&								& 			&			& Values		& Ratio		\\ \hline
	1  	& pima							& 768		& 8	 		& 2			& 1.87		\\ \hline
	2	& dermatology 						& 366		& 34			& 6			& 5.55		\\ \hline
	3	& segment0						& 2308		& 19			& 2			& 6.02		\\ \hline
	4	& led7digit 						& 443		& 7			& 2			& 10.97		\\ \hline
	5	& abalone9-18						& 731		& 8			& 2			& 16.4		\\ \hline
	6	& yeast							& 1484		& 8			& 10			& 23.15		\\ \hline
	7	& $poker-9_vs_7$					& 244		& 10			& 2			& 29.5		\\ \hline
	8	& kddcup-guess					& 1642		& 41			& 2			& 29.98		\\ 
		& passwd\_vs\_satan				&			&			&			&			\\ \hline
	9	& yeast5							& 1484		& 8			& 2			& 38.73		\\ \hline
	10	& ecoli							& 336		& 7			& 8			& 71.5		\\ \hline
	11	& abalone19						& 4174		& 8			& 2			& 129.44		\\ \hline
	12	& Page Blocks 						& 548		& 10			& 5			& 164		\\ \hline
	13	& Statlog (Shuttle)					& 2175		& 9			& 7			& 853		\\ \hline
  \end{tabular}
\end{table}

\subsection{Evaluation Methods}
ROC curves can be thought of as representing the family of best decision boundaries for relative costs of true positive (TP) and false positive (FP). On an ROC curve the X-axis represents and the Y-axis represents.

\begin{equation}
T P = { TP \over T P + F N  }
\end{equation}

\begin{equation}
F P = {  FP \over F P + T N  }
\end{equation}

The ideal point on the ROC curve would be (0,100), that is all positive instances are classified correctly and no negative instances are misclassified as positive. The line y = x represents the scenario of randomly guessing the class. Area Under the ROC Curve (AUC) is a useful metric for classifier performance as it is independent of the decision criterion selected and prior probabilities.The AUC comparison can establish a dominance relationship between classifiers. If the ROC curves are intersecting, the total AUC is an average comparison between models. However, for some specific cost and class distributions, the classifier having maximum AUC may in fact be suboptimal. Hence, we also compute the ROC convex hulls, since the points lying on the ROC convex hull are potentially optimal. 

\subsection{Results}
In this experiment, we have compared the proposed CUSBoost method with AdaBoost, RUSboost, and SMOTEBoost methods. Each dataset was validated using Area Under the ROC Curve (AUC). As for base learner we used C4.5 decision tree induction in boosting. Keel-dataset repository's implementation was used for the AdaBoost, RUSBoost, and SMOTEBoost algorithms \cite{KEEL-dataset}. Each of these experiments are done 5 times with 10 fold cross validation and their mean scores are shown in Table \ref{Mean_Result}.

\begin{table}[h!]\footnotesize
  \centering
  \caption{Average performance of the AdaBoost, RUSboost, SMOTEBoost, and CUSBoost methods on 13 imbalanced datasets.}\label{Mean_Result}
  \begin{tabular}{|l|l|l|l|l|l|}	\hline
        	Datasets				& AdaBoost		& RUSBoost	& SOMTEBoost	& CUSBoost  		\\ \hline
	pima					& 0.6223			& 0.6376	 	& 0.6597			& \textbf{0.6679}	\\ \hline
	dermatology 			& \textbf{0.9342}	& 0.43		& 0.632			& 0.54			\\ \hline
	segment0				& \textbf{0.996}		& 0.957		& 0.951			& 0.943			\\ \hline
	led7digit 				& 0.8937			& 0.8279		& 0.9373			& \textbf{0.9412}	\\ \hline
	abalone9-18			& 0.6934			& 0.7051		& 0.7195			& \textbf{0.7243}	\\ \hline
	yeast				& 0.7589			& 0.7382		& 0.741			& \textbf{0.7603}	\\ \hline
	$poker-9_vs_7$		& 0.642			& 0.589		& 0.6708			& \textbf{0.967}		\\ \hline
	kddcup-guess			& 0.8324			& 0.8681		& 0.8513			& \textbf{0.8745}	\\ 
	passwd\_vs\_satan		&				&			&				&				\\ \hline
	yeast5				& 0.9231			& 0.887		& \textbf{0.9253}	& 0.9				\\ \hline
	ecoli					& 0.6354			& 0.517		& \textbf{0.6597}	& 0.6589			\\ \hline
	abalone19				& 0.5723			& 0.5923		& 0.5583			& \textbf{0.609}		\\ \hline
	Page Blocks 			& 0.7605			& 0.5975		& 0.8123			& \textbf{0.8981}	\\ \hline
	Statlog (Shuttle)		& 0.8331			& 0.727		& 0.8445			& \textbf{0.8847}	\\ \hline
  \end{tabular}
\end{table}

From the results we can see that the CUSBoost algorithm performs best most of the time when the imbalance ratio is in specific range. As the imbalance ratio gets high CUSBoost starts to outperform all the other methods significantly. This happens because our method does not focus on making the ratio of majority and minority class examples $1:1$. So for this reason the sub sampled training dataset holds a better representation of the majority class while remaining imbalance itself. RUSBoost shows a performance with high variance especially with highly imbalance datasets, thus it shows a poor performance if the mean results are chosen. But, if the best results are chosen from 10 experiments then RUSBoost outperforms the other methods including the proposed method in many datasets those are shown in Table \ref{Best_Result}. The best result of proposed method is usually quite close to the average result thus proving low variance in its performance .          

\begin{table}[h!]\footnotesize
  \centering
  \caption{The best result using AdaBoost, RUSboost, SMOTEBoost, and CUSBoost methods in each dataset is stressed in bold-face.}\label{Best_Result}
  \begin{tabular}{|l|l|l|l|l|l|}	\hline
        	Datasets				& AdaBoost		& RUSBoost		& SOMTEBoost	& CUSBoost  		\\ \hline
	pima					& 0.6653			& \textbf{0.6796}	& 0.667			& 0.6679			\\ \hline
	dermatology 			& \textbf{0.9}		& 0.6935			& 0.632			& 0.54			\\ \hline
	segment0				& \textbf{0.998}		& 0.9805			& 0.9642			& 0.9523			\\ \hline
	led7digit 				& 0.9157			& 0.886			& 0.9485			& \textbf{0.9544}	\\ \hline
	abalone9-18			& 0.707			& \textbf{0.805}		& 0.7029			& 0.7434			\\ \hline
	yeast				& 0.7839			& \textbf{0.801}		& 0.7653			& 0.7895			\\ \hline
	$poker-9_vs_7$		& 0.657			& 0.7395			& 0.69			& \textbf{0.9685}	\\ \hline
	kddcup-guess			& 0.8578			& \textbf{0.9123}	& 0.8602			& 0.8849			\\ 
	passwd\_vs\_satan		&				&				&				&				\\ \hline
	yeast5				& 0.9324			& \textbf{0.9515}	& 0.9391			& 0.9126			\\ \hline
	ecoli					& 0.6455			& \textbf{0.677}		& 0.6753			& 0.6656			\\ \hline
	abalone19				& 0.5796			& \textbf{0.6498}	& 0.5876			& 0.6176			\\ \hline
	Page Blocks 			& 0.78			& \textbf{0.937}		& 0.8592			& 0.9007			\\ \hline
	Statlog (Shuttle)		& 0.8567			& \textbf{0.8989}	& 0.8516			& 0.8967			\\ \hline
  \end{tabular}
\end{table}

\section{Conclusion}
\label{conclusion}
Existing classification algorithms generally focus on majority class instances and ignore the minority class instances. So, it is a challenging task to construct an effective classifier that can correctly classify the instances of the minority class. This is ever so pertinent as class imbalance problems affect a vast range of domains. Recently, computational intelligence researchers have proposed several hybrid techniques by combining sampling with ensemble classifiers for dealing with class imbalance problems. The purpose of this paper is to present a new algorithm called CUSBoost, or Cluster-based Under-sampling with Boosting, in order to alleviate the problem of class imbalance. We have compared the performance of CUSBoost algorithm with the most effective boosting techniques like AdaBoost, RUSBoost, and SMOTEBoost algorithms. Based on experimental results, we have found that CUSBoost performed favourably when compared to these popular techniques on datasets having high class imbalance ratio. 

Rather than simply randomly choosing samples from the dataset, CUSBoost first clusters the majority class instances and then performs random under sampling so that the boosting algorithm (Adaboost) can use examples from all regions of the dataset. Thus an advantage it holds over RUSBoost is that the variance of the results are low, which leads to stable performance. Additionally, the performance of this algorithm has been shown to be relatively proportional to the imbalance ratio of the dataset, where higher imbalanced ratio produces greater performance. It has shown to outperform the other algorithms in a certain range of imbalanced ratio. In this paper, we have compared the performance of CUSBooost with RUSBoost, SMOTEBoost and Adaboost. In future work, we intend to perform extensive experiments to continue investigating the performance of CUSBoost with other established sampling methods and ensemble methods.


\bibliographystyle{IEEEtran}
\bibliography{Farid_CSITSS}

\end{document}